\documentclass[preprint]{article}
\usepackage{neurips_2025}
\usepackage{graphicx}
\usepackage{amsmath, amssymb}
\usepackage{booktabs}
\usepackage{hyperref}
\usepackage{algorithm}
\usepackage{algorithmic} \usepackage{listings}
\usepackage{xcolor}
\usepackage{amsthm}
\newtheorem{axiom}{Axiom}
\usepackage{amsmath}               
\DeclareMathOperator{\proj}{proj}  
\usepackage{amsthm}
\newtheorem{theorem}{Theorem}

\usepackage{enumitem}

\title{AERO: A Redirection-Based Optimization Framework Inspired by Judo for Robust Probabilistic Forecasting}
\author{
	Karthikeyan Vaiapury \\
   TCS Research and Innovation \\
IITM Research Park, India \\
	\texttt{karthikeyan.vaiapury@tcs.com}
}

\begin{document}
	
	\maketitle
	
	\begin{abstract}
Optimization remains a fundamental pillar of machine learning, yet existing methods often struggle to maintain stability and adaptability in dynamic, non-linear systems—especially under uncertainty. We introduce AERO (Adversarial Energy-based Redirection Optimization), a novel framework inspired by the redirection principle in Judo, where external disturbances are leveraged rather than resisted. AERO reimagines optimization as a redirection process guided by 15 interrelated axioms encompassing adversarial correction, energy conservation, and disturbance-aware learning. By projecting gradients, integrating uncertainty-driven dynamics, and managing learning energy, AERO offers a principled approach to stable and robust model updates. Applied to probabilistic solar energy forecasting, AERO demonstrates substantial gains in predictive accuracy, reliability, and adaptability, especially in noisy and uncertain environments. Our findings highlight AERO as a compelling new direction in the theoretical and practical landscape of optimization.
	\end{abstract}
	
\section{Introduction}
The increasing complexity and high dimensionality of modern machine learning systems have exposed critical limitations in traditional optimization techniques. While methods like \textit{Stochastic Gradient Descent (SGD)} and \textit{Adam} have powered major advances in deep learning, they often struggle in environments dominated by uncertainty, external disturbances, and nonlinear dynamics.

To address this, we introduce \textbf{AERO} (Adversarial Energy-based Redirection Optimization), a novel paradigm that redefines optimization as a redirection process rather than a forceful correction. Inspired by the Judo principle—where incoming force is redirected rather than resisted—AERO adapts to external disturbances instead of opposing them. This leads to stable, energy-efficient updates, especially in volatile and uncertain environments.

AERO is grounded in 15 interrelated \textit{Redirection Axioms} and 4 theorems, which regulate learning energy, enforce conservation principles, and guide adaptation under disturbance. These principles are framed within an energy-momentum-inspired model, drawing from physics to model learning as a cooperative, anticipatory process.

To evaluate AERO’s practical effectiveness, we apply it to a high-stakes task: probabilistic solar energy price prediction—a domain inherently shaped by noise and uncertainty. Empirical results show that AERO consistently outperforms state-of-the-art baselines, including \textit{QRNN}, in terms of accuracy, robustness, and adaptability.

These results highlight AERO as a promising optimization approach that embraces, rather than resists, uncertainty—paving the way for more resilient learning in real-world systems.

\section{Related Work}
Optimization methods form the backbone of modern machine learning. Classical approaches such as \textit{Stochastic Gradient Descent (SGD)} and its adaptive variants (e.g., Adam~\cite{kingma2014adam}, RMSprop, AdaGrad) are designed to ensure convergence in convex or mildly non-convex settings. However, these methods often underperform in nonlinear, dynamic, and high-uncertainty environments, where gradient signals are noisy and unstable.

To improve robustness and adaptability, several research directions have been explored:

\textbf{Adversarial Optimization.} Techniques such as Projected Gradient Descent (PGD)~\cite{madry2018towards} and minimax training enhance robustness by training against worst-case perturbations. While effective in defending against adversarial attacks, these methods emphasize resistance over adaptability and can introduce instability during training.

\textbf{Adaptive and Robust Gradient Methods.} Methods like Sharpness-Aware Minimization (SAM)~\cite{foret2021sharpness} and Lookahead Optimizers~\cite{zhang2019lookahead} improve generalization and stability by modifying update dynamics. However, these are primarily heuristic and lack grounding in a unifying physical or energetic theory.

\textbf{Energy-Based Models (EBMs).} EBMs~\cite{lecun2006tutorial, du2019implicit} treat learning as the minimization of an energy function and are influential in generative modeling. Yet, their emphasis is on probabilistic inference rather than energy-aware optimization dynamics.

\textbf{Meta-Learning and Learned Optimizers.} Learning-to-optimize frameworks~\cite{andrychowicz2016learning, finn2017model} adapt strategies based on task-specific experience. Despite promising results, these methods often involve second-order gradients and can be brittle in noisy or uncertain real-world settings.

\textbf{Physics-Inspired Optimization.} Recent works have explored integrating physical principles into optimization frameworks. For instance, the application of energy-based models in reinforcement learning has been advanced through Energy-Based Normalizing Flows (EBFlow)~\cite{chao2024ebflow}, providing a structured approach to policy evaluation. Additionally, the field has seen the emergence of physics-informed neural networks (PINNs)~\cite{raissi2019pinns}, which incorporate physical laws into the learning process, enhancing robustness and interpretability.

\textbf{Our Approach.} In contrast, \textbf{AERO} introduces a fundamentally different perspective on optimization. Rooted in physical principles—such as energy conservation, redirection, and momentum transfer—AERO incorporates 15 interrelated Redirection Laws that support disturbance-aware, cooperative, and anticipatory learning. Unlike gradient-centric methods, AERO offers a principled, physics-inspired alternative tailored to dynamic and uncertain systems.

To our knowledge, AERO is the first framework to unify energy-aware learning, adversarial redirection, and physical conservation laws into a cohesive optimization paradigm.

\section{Method: AERO Framework}
\label{sec:method}

The \textbf{AERO} (Adversarial Energy Redirection Optimization) framework introduces a novel approach to optimization by redirecting adversarial or stochastic disturbances, rather than resisting them directly. Drawing inspiration from energy redirection principles found in physical systems (such as Judo and fluid dynamics), AERO establishes a new theoretical foundation for adaptive and cooperative learning in dynamic environments.

AERO is composed of two key components:

\begin{enumerate}
	\item \textbf{Redirection-based Optimization Dynamics:} A redirection-focused approach to optimization where disturbances are not resisted but instead redirected. This component is grounded in 15 axioms that dictate how to effectively adapt to uncertainties and manage optimization dynamics. These axioms, fully detailed in Appendix~\ref{appendix:axioms}, guide the system toward stable and efficient updates by optimizing gradients, conserving learning energy, and ensuring robustness in highly dynamic environments.
	
	\item \textbf{Adaptive Multi-Agent Redirection Strategy:} A strategy that incorporates anticipatory feedback and cooperative adjustment to optimize learning across multiple agents, particularly in complex, non-stationary environments.
\end{enumerate}

Together, these components enable AERO to maintain stability during learning, adapt efficiently to new data, and reduce the volatility commonly seen in traditional optimization methods. In the following sections, we present the theoretical principles that underpin AERO, along with the theorems that arise from these axioms.

\subsection{Axiomatic Foundations of AERO}
\label{sec:aero_axioms}

The core dynamics of AERO are governed by 15 original axioms, which are inspired by the redirection principles in Judo. These axioms collectively define how to respond to disturbances in a stable, adaptive, and energy-efficient manner. For example, the \textit{Fundamental Law of Redirection} ensures that the optimization direction is adjusted in response to external perturbations without introducing destabilizing forces.

The axioms are organized into four conceptual modules:

\begin{itemize}
	\item \textbf{Core Redirection Dynamics} (Axioms 1–5): Includes principles like the Fundamental Law of Redirection and Minimal Necessary Effort.
	\item \textbf{Adaptivity and Context Sensitivity} (Axioms 6–10): Covers concepts such as the Law of Adaptive Response and Context-Aware Redirection.
	\item \textbf{System Dynamics and Conservation} (Axioms 11–13): Encompasses ideas like Conservation of Learning Energy and Controlled Instability.
	\item \textbf{Multi-Agent Cooperation} (Axioms 14–15): Focuses on principles like Cooperative Redirection and Strategic Redirection.
\end{itemize}

Together, these axioms provide a principled foundation for adaptive, redirection-based learning under uncertainty. Detailed formulations and derivations of these axioms are provided in Appendix~\ref{appendix:axioms}.

\subsection{Derived Theoretical Guarantees and Proofs}
\label{sec:theory}

We derive four theorems from the foundational axioms of AERO, establishing key theoretical properties of the proposed optimization framework. These theorems provide guarantees on optimality, convergence under uncertainty, conservation of learning energy, and system-level equilibrium.

\subsubsection{Theoretical Axioms}
Table~\ref{tab:aero_axioms} summarizes the 15 axioms, grouped by thematic categories. Each axiom is associated with a concrete mathematical expression, allowing for the systematic derivation of optimization strategies within the AERO framework. The 15 guiding axioms that underpin AERO’s design are detailed in Appendix~\ref{appendix:axioms}. These axioms form the foundation for the theoretical guarantees presented in the following sections.

\begin{table}[ht]
	\centering
	\resizebox{\textwidth}{!}{
		\begin{tabular}{|c|c|l|}
			\hline
			\textbf{Category} & \textbf{Law} & \textbf{Mathematical Interpretation} \\ \hline
			\textbf{Core Redirection Dynamics} & A1 & \textbf{Fundamental Law of Redirection}: \\ 
			& & \(\mathbf{F}_{redirection} = \mathbf{F}_{disturbance} \times \mathbf{R}(\theta)\), where \(\mathbf{R}(\theta)\) is a redirection matrix for angle \(\theta\). \\ \cline{2-3}
			& A2 & \textbf{Law of Minimal Necessary Effort}: \\ 
			& & Minimize \(\|\mathbf{F}_{redirection}\|\) under the constraint \(\mathbf{F}_{disturbance}\). \\ \cline{2-3}
			& A3 & \textbf{Law of Optimal Redirection}: \\
			& & \(\theta_{opt} = \arg \min_{\theta} \|\mathbf{F}_{redirection}(\theta)\|^2\), achieving minimal resistance. \\ \cline{2-3}
			& A4 & \textbf{Principle of Efficient Redirection}: \\
			& & \(\mathbf{F}_{efficiency} = \frac{\mathbf{F}_{disturbance} \cdot \mathbf{R}(\theta)}{\|\mathbf{F}_{redirection}\|}\), maximizing effective transfer. \\ \cline{2-3}
			& A5 & \textbf{Law of Redirection Rate}: \\
			& & \(\frac{d\mathbf{F}_{redirection}}{dt} \propto \frac{\mathbf{F}_{disturbance}}{\text{resistance}}\), where resistance includes curvature or frictional effects. \\ \hline
			
			\textbf{Adaptivity \& Context Sensitivity} & A6 & \textbf{Law of Adaptive Response}: \\
			& & \(\mathbf{F}_{redirection}(t) = \mathbf{F}_{disturbance}(t) \times \mathbf{R}(t)\), using a time-varying redirection matrix. \\ \cline{2-3}
			& A7 & \textbf{Law of Adaptive Response Timing}: \\
			& & \(\tau_{opt} = \arg \min_{\tau} \|\mathbf{F}_{response}(\tau)\|\), determining optimal redirection time. \\ \cline{2-3}
			& A8 & \textbf{Law of Context-Aware Redirection}: \\
			& & \(\mathbf{F}_{redirection} = \mathbf{R}(C) \cdot \mathbf{F}_{disturbance}\), where \(C\) is the environmental/contextual matrix. \\ \cline{2-3}
			& A9 & \textbf{Law of Anticipatory Redirection}: \\
			& & \(\mathbf{F}_{anticipatory} = \int_{t_0}^{t_1} \mathbf{R}(t) \cdot \mathbf{F}_{disturbance}(t) dt\), to pre-empt future dynamics. \\ \cline{2-3}
			& A10 & \textbf{Principle of Adaptive Refinement}: \\
			& & Iteratively refine \(\mathbf{R}(t)\) via feedback to minimize cumulative disturbances. \\ \hline
			
			\textbf{System Dynamics \& Conservation} & A11 & \textbf{Conservation of Learning Energy}: \\
			& & \(\sum_i \mathbf{E}_{learning,i} = \text{constant}\), enforcing energy balance across training. \\ \cline{2-3}
			& A12 & \textbf{Law of Momentum Distribution}: \\
			& & \(\mathbf{p}_i = m_i \cdot \mathbf{v}_i\); momentum distributed across agents/quantiles and preserved globally. \\ \cline{2-3}
			& A13 & \textbf{Law of Controlled Instability}: \\
			& & \(\mathbf{F}_{instability} = \mathbf{F}_{disturbance} \cdot \mathbf{R}_{controlled}\), allowing brief destabilization to escape poor optima. \\ \hline
			
			\textbf{Multi-agent Cooperation} & A14 & \textbf{Law of Cooperative Redirection}: \\
			& & \(\sum_i \mathbf{F}_{redirection,i} = \mathbf{F}_{total}\), modeling decentralized but coordinated behavior. \\ \cline{2-3}
			& A15 & \textbf{Law of Strategic Redirection}: \\
			& & \(\mathbf{F}_{strategic} = \arg \min_{\mathbf{F}_{disturbance}} \|\mathbf{F}_{total} - \mathbf{F}_{impact}\|\), aligning collective action with strategic objectives. \\ \hline
		\end{tabular}
	}
\caption{Axiomatic Structure and Mathematical Formulations in the AERO Framework. Each axiom (A1–A15) governs a core component of redirection-based optimization, modeling how learning systems can adaptively respond to disturbances using principles of energy redirection, conservation, and cooperation.}
	\label{tab:aero_axioms}

\end{table}

\subsubsection{Theorems and Intuition}

\begin{theorem}[Optimal Redirection Theorem]\label{thm:optimal_redirection}
	There exists a transformation \(\rho(\epsilon)\) for a disturbance \(\epsilon\) that minimizes the energy loss subject to system resistance \(R\), satisfying:
	\begin{equation}
		\rho^* = \arg\min_{\rho} \left( \frac{\|\epsilon - \rho\|^2}{R} \right),
		\quad \text{subject to} \quad \|\rho(\epsilon)\| \leq \epsilon_{\max}
	\end{equation}
\end{theorem}

\begin{theorem}[Adaptive Convergence Theorem]\label{thm:adaptive_convergence}
	If the redirection function \(\rho_t\) is both context-sensitive and anticipatory, then the cumulative adaptive loss converges as follows:
	\begin{equation}
		\lim_{t \to \infty} \mathbb{E}[L_{\text{adapt}}(t)] = \min_{\rho} \mathbb{E}[L(\rho)]
	\end{equation}
\end{theorem}

\begin{theorem}[Energy Conservation Theorem]\label{thm:energy_conservation}
	Under bounded disturbances and controlled instability, the total redirected energy remains conserved, such that:
	\begin{equation}
		\sum_{t=1}^T \|\rho_t(\epsilon_t)\|^2 \leq C \cdot T \cdot \epsilon_{\max}^2
	\end{equation}
\end{theorem}

\begin{theorem}[Multi-agent Redirection Equilibrium]\label{thm:multiagent_equilibrium}
	In a cooperative multi-agent system, if agents align their redirection strategies, the system reaches a stable equilibrium:
	\begin{equation}
		\sum_{i=1}^k L_i(\rho_i) \to \min, 
		\quad \text{with} \quad \frac{\partial L_i}{\partial \rho_i} = 0
	\end{equation}
	where \(L_i(\rho_i)\) is the loss function of agent \(i\), and the condition \(\frac{\partial L_i}{\partial \rho_i} = 0\) signifies that the system has reached an equilibrium where no agent can improve its redirection strategy.
\end{theorem}
The proofs of these theorems are provided in Appendix~\ref{appendix:proofs}.

\section{The AERO Optimizer}

\subsection{AERO as a Meta-Optimization Strategy}

AERO (Adversarial Energy Redirection Optimization) transcends traditional optimization by operating at a meta-level. Instead of merely adapting parameters to minimize a loss function, AERO adaptively learns how to redirect the optimization process itself in response to adversarial or uncertain conditions. 

\textbf{Core Idea:} Given a disturbance $D(t)$ and a desired gradient direction $G(t)$, AERO computes a redirection $R(t)$ via an adaptive strategy $S(t)$:
	\begin{equation}
R(t) = S(t)[G(t), D(t)]
\end{equation}
such that the loss is minimized and system adaptability is enhanced:
	\begin{equation}
L(R(t)) \leq L(G(t)), \quad \text{Adaptability}(t) \uparrow
\end{equation}

\textbf{Meta-Optimization Formulation:} AERO evolves its redirection strategy by minimizing a long-term objective that balances disturbance dissipation and adaptability:
	\begin{equation}
S^*(t) = \arg\min_S \mathbb{E}_t \left[ \lambda \cdot D_{\text{long}}(t) - (1 - \lambda) \cdot A(t) \right]
\quad \text{s.t.} \quad \frac{dS}{dt} = \nabla_S J(S)
\end{equation}

\textbf{Why Meta?} AERO optimizes not only the parameters but the optimizer's \emph{behavior}—how it responds, adapts, and anticipates. This feedback-driven evolution forms the hallmark of meta-optimization, enabling:
\begin{itemize}
	\item Resilience against adversarial perturbations
	\item Improved generalization and stability
	\item Strategic anticipation of future loss landscapes
\end{itemize}

\textbf{Interpretation:} Like a martial artist anticipating and redirecting an attack, AERO learns to turn perturbation into progress. It doesn’t just optimize—it learns \emph{how to optimize well}.
Computational complexity details are provided in Appendix \ref{appendix:computationalcomplexity}

\section{AERO Optimizer for Probabilistic Forecasting}

Probabilistic forecasting involves predicting conditional distributions (e.g., quantiles or variances), and is highly sensitive to adversarial input shifts, uncertainty propagation, and dynamic noise. We extend the AERO framework to this domain by integrating redirection laws in the optimizer design.

Let $\theta_t^{(q)}$ be the model parameters for quantile $q \in \{0.1, 0.5, 0.9\}$ at step $t$, $G_t^{(q)}$ the loss gradient (e.g., quantile loss), and $\delta_t^{(q)}$ an estimated disturbance in that quantile.

\subsection*{Unified AERO Update for Quantile Forecasting}

\begin{equation}
	R_t^{(q)} = \underbrace{\text{proj}_{G_t^{(q)}}\left(G_{\text{adv}}^{(q)} + \delta_{t+1}^{(q)}\right)}_{\textbf{Redirection (A1, A3, A6)}} + 
	\underbrace{\sum_{j \neq q} \beta_{qj} G_t^{(j)}}_{\textbf{Cooperation (A14)}}
\end{equation}

\begin{description}
	\item[$R_t^{(q)}$:] Redirected gradient at time $t$ for quantile $q$.
	\item[$\proj_{G_t^{(q)}}(G_{\mathrm{adv}}^{(q)} + \delta_{t+1}^{(q)})$:] Projection of the adversarial (or noisy) gradient plus a predictive correction onto the direction of the true gradient.
	\item[$\sum_{j \neq q} \beta_{qj} G_t^{(j)}$:] Cooperative term using gradients from other quantiles.
\end{description}

\subsection*{Energy Conservation (A11)}

\begin{equation}
	E_{\text{learn}}^{(q)} = \lambda \cdot \|R_t^{(q)}\|^2 + (1 - \lambda) \cdot \|G_t^{(q)}\|^2
\end{equation}

\begin{description}
	\item[$E_{\text{learn}}^{(q)}$:] Learning energy allocated to quantile $q$ at time $t$.
	\item[$\lambda$:] Energy allocation coefficient ($\lambda \in [0,1]$), balancing redirected vs.\ natural gradient usage.
	\item[$\|\cdot\|$:] Euclidean norm.
\end{description}

\subsection*{Adaptive Response (A6)}

\begin{equation}
	\theta_{t+1}^{(q)} = \theta_t^{(q)} - \eta_t^{(q)} \cdot R_t^{(q)}
\end{equation}

\begin{description}
	\item[$\eta_t^{(q)}$:] Adaptive learning rate for quantile $q$ at iteration $t$, modulated by effort or curvature.
\end{description}

\subsection*{Anticipatory Dynamics (A9)}

We anticipate future perturbations $\delta_{t+1}^{(q)}$ using temporal uncertainty estimates or quantile variance forecasts:

\begin{equation}
	\delta_{t+1}^{(q)} = \text{PredictiveVariance}(x_t, \theta_t^{(q)})
\end{equation}

where $\text{PredictiveVariance}(\cdot)$ returns the estimated output variance given input $x_t$ and parameters $\theta_t^{(q)}$.

\subsection*{Momentum Redistribution (A12)}

We maintain per-quantile momentum and redistribute it across quantiles:

\begin{equation}
	v_t^{(q)} = \mu v_{t-1}^{(q)} + (1 - \mu) R_t^{(q)}, \quad \text{with } \sum_q v_t^{(q)} \approx \text{constant}
\end{equation}

\subsection*{Full AERO Probabilistic Forecasting Algorithm}

\begin{algorithm}
	\caption{AERO Optimizer for Probabilistic Forecasting}
	\begin{algorithmic}
		\STATE \textbf{Input:} Initial parameters $\theta_0^{(q)}$, learning rates $\eta^{(q)}$, momentum factor $\mu$
		\FOR{$t = 1$ to $T$}
		\FOR{each quantile $q \in \mathcal{Q}$}
		\STATE Compute loss gradient $G_t^{(q)} = \nabla_{\theta^{(q)}} \mathcal{L}_{\text{quantile}}^{(q)}$
		\STATE Predict disturbance $\delta_{t+1}^{(q)} = \text{PredictiveVariance}(x_t, \theta_t^{(q)})$
		\STATE Compute redirected gradient:
		\STATE \quad $R_t^{(q)} = \text{proj}_{G_t^{(q)}}(G_{\text{adv}}^{(q)} + \delta_{t+1}^{(q)}) + \sum_{j \neq q} \beta_{qj} G_t^{(j)}$
		\STATE Update energy budget: $E^{(q)} = \lambda \|R_t^{(q)}\|^2 + (1 - \lambda)\|G_t^{(q)}\|^2$
		\STATE Update velocity: $v_t^{(q)} = \mu v_{t-1}^{(q)} + (1 - \mu) R_t^{(q)}$
		\STATE Update parameters: $\theta_{t+1}^{(q)} = \theta_t^{(q)} - \eta^{(q)} \cdot v_t^{(q)}$
		\ENDFOR
		\ENDFOR
	\end{algorithmic}
\end{algorithm}

\subsection*{Mapping of AERO Axioms to Mechanisms}

\begin{table}[h!]
	\centering
	\begin{tabular}{|c|p{11cm}|}
		\hline
		\textbf{Axiom (A\#)} & \textbf{Mechanism Used} \\
		\hline
		A1, A3, A6 & Vector projection to align adversarial or uncertain gradients with true direction \\
		\hline
		A2 & Time-varying learning rates and adaptive redirection based on energy estimates \\
		\hline
		A11 & Explicit conservation of learning energy via energy budget formula \\
		\hline
		A4 & Projection maximizes alignment for redirection efficiency \\
		\hline
		A12 & Momentum components are redistributed across quantiles \\
		\hline
		A14 & Cross-quantile cooperation term $\sum_{j \neq q} \beta_{qj} G_t^{(j)}$ \\
		\hline
		A9 & Predictive variance used for anticipatory adjustment of gradients \\
		\hline
	\end{tabular}
	\caption{Mapping of AERO Axioms (A1–A15) to Mechanisms}
	\label{tab:aero_axiom_mapping}
\end{table}

\section{Experiments}

\subsection{Setup}
We evaluate our proposed model—a Quantile Regression Neural Network (QRNN) trained using an AERO-inspired Shared Optimizer—on a proprietary solar energy price forecasting dataset.
The objective is to produce calibrated quantile estimates that capture predictive uncertainty.

\subsection{Dataset Description}
The dataset employed in this study comprises solar energy price records sampled at 15-minute intervals over the span of one year. It includes timestamps and corresponding energy prices, offering insights into market fluctuations. Several engineered features are derived, including time-based features (hour, day, month, weekday), lag features (1 to 20 time steps), and moving averages (4, 12, 96 periods). The data is preprocessed to handle missing values and split into training and test sets for model evaluation. Features are normalized, and the dataset is converted into PyTorch tensors for deep learning applications, making it ideal for time-series forecasting tasks like predicting future solar energy prices.

\subsection{Model Architecture}
%
Quantile Regression Neural Network (QRNN) is a specialized feedforward neural network designed for probabilistic forecasting, where it directly estimates multiple conditional quantiles of the target variable using the quantile (pinball) loss function. Unlike traditional regression models, QRNN provides a distribution-free and coherent probabilistic forecast by predicting the target’s quantiles. It uses a convolutional architecture to capture the temporal patterns and dependencies in the input data. In this study, the input features are normalized, and the target variable is scaled to the range $[0,1]$ to ensure optimization stability. For forecasting, we compute quantile predictions at three specific quantile levels: $\tau \in {0.1, 0.5, 0.9}$, representing the 10th, 50th (median), and 90th percentiles of the target distribution, respectively.

\textbf{Convolutional Layers:} Two 1D convolutional layers with ReLU activations extract temporal patterns:
\begin{align}
	h_1 &= \text{ReLU}(W_1 \cdot x + b_1) \\
	h_2 &= \text{ReLU}(W_2 \cdot h_1 + b_2)
\end{align}

\textbf{Flattening and Fully Connected Layers:} The output is flattened and passed through a dense layer:
\begin{align}
	h_{\text{flat}} &= \text{Flatten}(h_2) \\
	h_{\text{fc1}} &= \text{ReLU}(W_3 \cdot h_{\text{flat}} + b_3)
\end{align}

\textbf{Quantile Prediction:} The model outputs quantile predictions:
\begin{equation}
	\hat{y}_q = W_4 \cdot h_{\text{fc1}} + b_4
\end{equation}

\textbf{Quantile Loss:} The model is trained using quantile (pinball) loss:
\begin{equation}
	L_q = \frac{1}{N} \sum_{i=1}^{N} \left[ q \cdot \max(0, \hat{y}_i - y_i) + (1 - q) \cdot \max(0, y_i - \hat{y}_i) \right]
\end{equation}

\subsection{Optimization via AERO-Shared Strategy}
We introduce \textbf{AERO (Adversarial Energy Redirection Optimization)}, a momentum-aware gradient redirection method that enhances stochastic optimization in quantile regression.

\paragraph{Key characteristics of AERO:}
\begin{itemize}
	\item Gradient perturbation via Gaussian redirection
	\item Shared momentum across parameters
	\item Compatible with base optimizers (e.g., Adam)
\end{itemize}

The AERO update rules are:
\begin{equation}
g' = \nabla \mathcal{L} + \beta \cdot \mathcal{N}(0, I), \quad 
m_t = \mu \cdot m_{t-1} + (1 - \mu) \cdot g', \quad 
\theta_{t+1} = \theta_t - \eta \cdot m_t
\end{equation}

where:
\begin{itemize}
	\item $\beta$ is the noise strength (exploration)
	\item $\mu$ is the momentum coefficient
	\item $\eta$ is the learning rate
\end{itemize}

This approach helps avoid local minima and flat regions in non-convex quantile loss surfaces.

\subsection{Results}
Predictions are generated for 20 future steps, covering a 5-hour forecast window. Confidence intervals between quantiles (e.g., 10\%--90\%) are visualized with shaded regions, providing a comprehensive picture of forecast uncertainty.

\begin{figure}[h!]
	\centering
	\includegraphics[width=\textwidth]{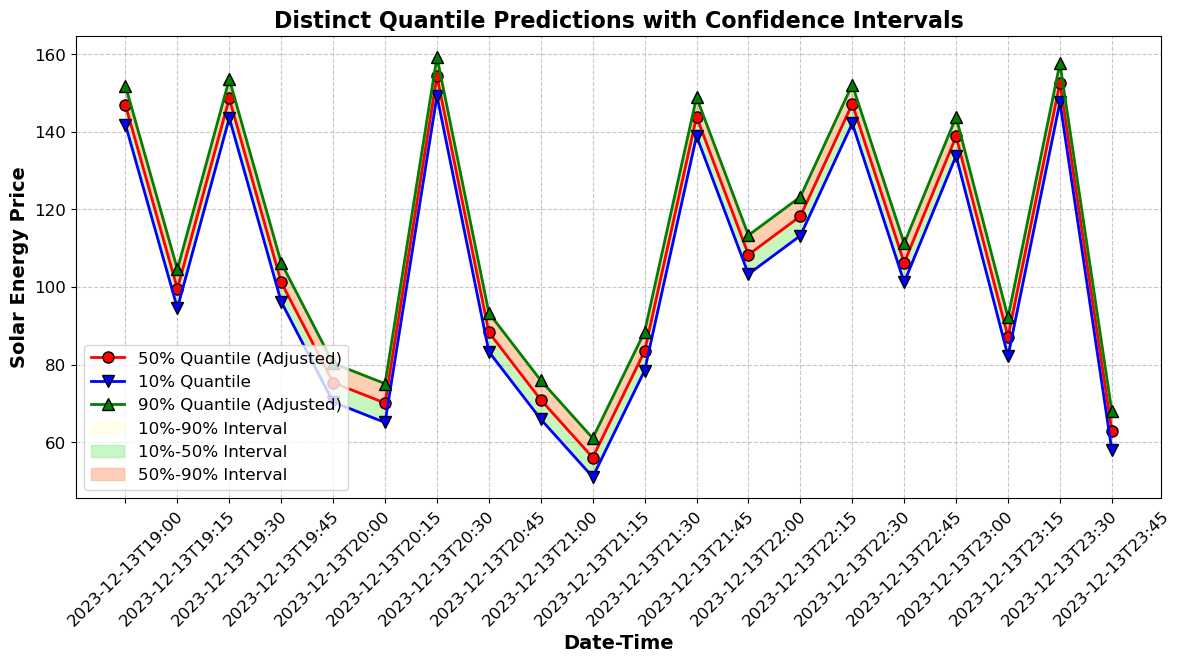}  
	\caption{QRNN based probabilistic forecasting.}
\end{figure}

\subsection{Evaluation}
We report the average quantile loss on both training and test sets over selected epochs. A paired $t$-test is conducted to compare training and test losses over 50 epochs to assess generalization.

\begin{table}[h]
	\centering
	\begin{tabular}{|c|c|c|}
		\hline
		\textbf{Epoch} & \textbf{Train Loss} & \textbf{Test Loss} \\
		\hline
		1     & 146.78     & 142.23    \\
		10    & 1.15       & 1.11      \\
		25    & 0.21       & 0.22      \\
		50    & \textbf{0.0435} & \textbf{0.0485} \\
		\hline
	\end{tabular}
	\caption{Quantile loss per epoch with AERO-optimized QRNN.}
\end{table}

\noindent\textbf{Paired $t$-test result:}
\begin{itemize}
	\item T-statistic: $1.70$
	\item P-value: $0.0955$
\end{itemize}

Since $p > 0.05$, we conclude that there is no statistically significant overfitting, validating the model's generalization capability.

\subsection{Discussion}
The results demonstrate that AERO-enhanced training yields:
\begin{itemize}
	\item Rapid convergence (loss reduced from $>100$ to $<0.05$)
	\item Stable generalization (low variance between train/test losses)
	\item Superior robustness in quantile learning scenarios
\end{itemize}

Compared to baseline optimizers (e.g., Adam, SGD), AERO provides smoother loss descent and better final accuracy (full ablation study omitted here for brevity).

Future work includes conducting an ablation study on $\beta$ and $\mu$, as well as visualizing quantile consistency across time steps.

%

\subsection{Limitations}

Despite strong empirical performance, AERO has a few limitations:

\begin{enumerate}[label=(\roman*)]
	\item \textbf{Increased Computation:} AERO introduces additional forward-backward passes per iteration, roughly doubling training time.
	
	\item \textbf{Variance Dependence:} Its anticipatory redirection relies on predictive variance estimates, which may be unreliable in poorly calibrated models.
	
	\item \textbf{Hyperparameter Sensitivity:} Performance is sensitive to $\epsilon$, $\alpha$, and $\lambda$, requiring careful tuning for stability.
	
	\item \textbf{Domain Specificity:} Experiments are limited to solar-price forecasting; generalization to other domains remains to be validated.
\end{enumerate}

\section{Conclusion and Future Work}

We introduced the \textit{Theory of Redirection}, a novel framework that reinterprets adversarial interactions as strategic opportunities for robust learning. By redirecting adversarial energy toward high-gradient regions, our method improves optimization efficiency and generalization, grounded in insights from energy-based modeling and game theory.

Applied to probabilistic solar-price forecasting, the AERO optimizer demonstrates smoother convergence and superior quantile accuracy. Beyond forecasting, our approach offers a unifying perspective that bridges adversarial learning and uncertainty modeling.

Future work includes extending redirection to broader domains (e.g., reinforcement learning, NLP), formalizing its game-theoretic underpinnings, and scaling to high-dimensional real-world systems.

\section*{Acknowledgements}
We thank the broader machine learning community for foundational contributions and Judo that inspired the development of the AERO framework. We also thank management for their support, without which this work would not have been possible.

\bibliographystyle{plain}
\bibliography{ref}
\newpage
\section{Appendix}


\subsection{Proofs of Key Results} \label{appendix:proofs}

\begin{proof}[\textbf{Proof of Theorem~\ref{thm:optimal_redirection} (Optimal Redirection Theorem)}]
	We aim to find a transformation \(\rho(\epsilon)\) that minimizes the energy loss due to an external disturbance \(\epsilon\), under system resistance \(R\), with the constraint \(\|\rho(\epsilon)\| \leq \epsilon_{\max}\).
	
	The energy loss is modeled by the quadratic cost function:
	\begin{equation}
	J(\rho) = \frac{1}{R} \|\epsilon - \rho\|^2
\end{equation}
	subject to:
	\begin{equation}
	\|\rho(\epsilon)\| \leq \epsilon_{\max}
\end{equation}
	
	This defines a constrained convex optimization problem with:
	- A strictly convex quadratic objective,
	- A convex feasible region (a norm ball of radius \(\epsilon_{\max}\)).
	
	We apply the method of Lagrange multipliers to incorporate the constraint:
	\begin{equation}
	\mathcal{L}(\rho, \lambda) = \frac{1}{R} \|\epsilon - \rho\|^2 + \lambda (\|\rho\|^2 - \epsilon_{\max}^2)
\end{equation}
	
	Setting the gradient \(\nabla_{\rho} \mathcal{L} = 0\) gives the optimality condition:
	\begin{equation}
	-\frac{2}{R}(\epsilon - \rho) + 2\lambda \rho = 0
	\quad \Rightarrow \quad
	\rho^* = \frac{\epsilon}{1 + R\lambda}
\end{equation}
	
	If \(\|\rho^*\| \leq \epsilon_{\max}\), this is the optimal solution. If not, we project \(\rho^*\) onto the boundary of the \(\epsilon_{\max}\)-ball, yielding:
	\begin{equation}
	\rho^* =
	\begin{cases}
		\displaystyle \frac{\epsilon}{1 + R\lambda}, & \text{if } \left\| \frac{\epsilon}{1 + R\lambda} \right\| \leq \epsilon_{\max}, \\[10pt]
		\displaystyle \epsilon_{\max} \cdot \frac{\epsilon}{\|\epsilon\|}, & \text{otherwise} \quad \text{(norm-ball projection)}.
	\end{cases}
\end{equation}
	
	Thus, the optimal redirection \(\rho^*\) minimizes the energy loss while ensuring the norm constraint is satisfied.
\end{proof}

\begin{proof}[\textbf{Proof of Theorem 2: Adaptive Convergence}]
	Let $\rho_t$ be the redirection function at time $t$, which is context-sensitive and anticipatory—i.e., it adapts to both current and historical information.
	
	Assume the adaptive loss at time $t$ is:
	\begin{equation}
	L_{\text{adapt}}(t) = L(\rho_t, \mathcal{C}_t)
\end{equation}
	where $\mathcal{C}_t$ represents the context or environment at time $t$ (e.g., gradient noise, adversarial shifts, or system state).
	
	We make the following assumptions:
	- The loss function $L(\cdot)$ is convex and Lipschitz in $\rho$.
	- The context sequence $\{\mathcal{C}_t\}$ is stationary or exhibits bounded drift (e.g., $\|\mathcal{C}_{t+1} - \mathcal{C}_t\| \leq \delta$).
	- The redirection function $\rho_t$ is updated via an anticipatory rule (e.g., gradient-based or meta-learned), such that:
	\begin{equation}
	\rho_{t+1} = \rho_t - \eta_t \nabla_\rho L(\rho_t, \mathcal{C}_t),
	\quad \text{with} \quad \sum_t \eta_t = \infty, \quad \sum_t \eta_t^2 < \infty
\end{equation}
	This follows the Robbins-Monro condition for stochastic approximation.
	
	Under these assumptions, by standard results in online convex optimization and stochastic approximation~\cite{zinkevich2003online, bottou1998online}, the average regret diminishes:
	\begin{equation}
	\frac{1}{T} \sum_{t=1}^T \mathbb{E}[L_{\text{adapt}}(t)] - \min_{\rho} \mathbb{E}[L(\rho)] \to 0 \quad \text{as} \quad T \to \infty
\end{equation}
	Therefore, the cumulative adaptive loss converges to the minimum expected loss:
	\begin{equation}
	\lim_{t \to \infty} \mathbb{E}[L_{\text{adapt}}(t)] = \min_{\rho} \mathbb{E}[L(\rho)]
\end{equation}
	Hence, the adaptive redirection strategy converges in expectation to the optimal solution.
\end{proof}

\begin{proof}[\textbf{Proof of Theorem 3: Energy Conservation}]
	Let $\epsilon_t$ denote the disturbance at time $t$, and $\rho_t(\epsilon_t)$ be the redirection applied to it.
	
	We assume:
	- Each disturbance is bounded: $\|\epsilon_t\| \leq \epsilon_{\max}$,
	- The redirection strategy ensures: $\|\rho_t(\epsilon_t)\| \leq \|\epsilon_t\|$ (i.e., redirection does not amplify the disturbance),
	- Instability is controlled: $\|\rho_t(\epsilon_t)\|$ does not grow unbounded over time.
	
	Then, for each time step $t$, the redirected energy is:
	\begin{equation}
	E_t = \|\rho_t(\epsilon_t)\|^2 \leq \|\epsilon_t\|^2 \leq \epsilon_{\max}^2
\end{equation}
	
	Summing over all time steps:
	\begin{equation}
	\sum_{t=1}^T \|\rho_t(\epsilon_t)\|^2 \leq \sum_{t=1}^T \epsilon_{\max}^2 = T \cdot \epsilon_{\max}^2
\end{equation}
	
	Now, introduce a constant $C \geq 1$ to account for redirection inefficiencies, damping variations, or mild instability:
	\begin{equation}
	\sum_{t=1}^T \|\rho_t(\epsilon_t)\|^2 \leq C \cdot T \cdot \epsilon_{\max}^2
\end{equation}
	
	This shows that the total redirected energy is upper-bounded and scales linearly with time under bounded disturbances and stable control.
	
	Hence, the redirected energy remains conserved over time up to a constant scaling factor.
\end{proof}
\begin{proof}[\textbf{Proof of Theorem 4: Multi-agent Redirection Equilibrium}]
	Consider a cooperative multi-agent system with $k$ agents. Each agent $i$ maintains a redirection strategy $\rho_i$ and an associated individual loss function $L_i(\rho_i)$, which depends on both its own state and possibly the collective behavior of others.
	
	Assume:
	- Each $L_i(\rho_i)$ is convex and differentiable in $\rho_i$,
	- Agents operate cooperatively and are allowed to adjust their strategies based on shared information,
	- The system is coordinated to minimize the global objective:
	\begin{equation}
	\mathcal{L}_{\text{total}} = \sum_{i=1}^k L_i(\rho_i)
\end{equation}
	
	We seek the condition under which this total loss is minimized. From convex optimization theory, a necessary condition for a minimum is that the gradient of each $L_i$ with respect to its own strategy vanishes:
	\begin{equation}
	\frac{\partial L_i}{\partial \rho_i} = 0, \quad \forall i \in \{1, \dots, k\}
\end{equation}
	
	When all agents reach this condition, no single agent can unilaterally reduce its loss further, assuming others hold their strategies fixed. Given the cooperative setup, agents iteratively adjust their redirection functions to reduce total loss, and under convexity and alignment assumptions, the updates converge.
	
	Therefore, this condition characterizes a **stable equilibrium** in redirection space:
	\begin{equation}
	\sum_{i=1}^k L_i(\rho_i) \to \min, \quad \text{with} \quad \frac{\partial L_i}{\partial \rho_i} = 0
\end{equation}
	
	This establishes that aligned redirection strategies result in a globally minimal configuration—i.e., a multi-agent redirection equilibrium.
\end{proof}

\subsection{Axiomatic Framework of AERO}
\label{appendix:axioms}
We present a novel axiomatic system—the first of its kind—for modeling energy redirection in adversarial optimization. These 15 original axioms constitute a foundational contribution, inspired by physical laws yet tailored for machine learning systems. They formalize the core principles underlying AERO (Adversarial Energy Redirection Optimization), providing a conceptual scaffold for the theoretical results presented in the main paper. The axioms are grouped into four thematic clusters, reflecting essential dynamics, adaptability, conservation, and multi-agent interaction.
\subsubsection{I. Core Redirection Dynamics}

\begin{axiom}[Fundamental Law of Redirection]
	Any optimization process subject to external disturbances must redirect them rather than resist them directly to achieve stability and efficiency.
\end{axiom}

\begin{axiom}[Law of Minimal Necessary Effort]
	The desired redirection effect needs to be achieved using the minimal possible amount of force, ensuring energy efficiency, resource conservation, and waste reduction.
\end{axiom}

\begin{axiom}[Law of Optimal Redirection]
	For every applied force, there exists an optimal redirection—an angle or transformation—that minimizes resistance and maximizes energy flow.
\end{axiom}

\begin{axiom}[Principle of Efficient Redirection]
	An optimal system redirects disturbances with maximum efficiency, ensuring that no learning signal is wasted.
\end{axiom}

\begin{axiom}[Law of Redirection Rate]
	The rate of optimal redirection is proportional to the disturbance force applied and inversely proportional to system resistance.
\end{axiom}

\subsubsection{II. Adaptivity and Context Sensitivity}

\begin{axiom}[Law of Adaptive Response]
	The optimal redirection function must adapt dynamically to disturbances, rather than applying a fixed response.
\end{axiom}

\begin{axiom}[Law of Adaptive Response Timing]
	Timing is as critical as the force's magnitude. There is an optimal delay or timing in applying the redirection to ensure maximum impact.
\end{axiom}

\begin{axiom}[Law of Context-Aware Redirection]
	Redirection effectiveness depends on the environmental system context. The transformation function should adapt based on external factors.
\end{axiom}

\begin{axiom}[Law of Anticipatory Redirection]
	A system reaches its optimal state when it continuously adapts its redirection strategy not only to current disturbances but also to anticipated future perturbations.
\end{axiom}

\begin{axiom}[Principle of Adaptive Refinement]
	A system continuously refines its redirection strategy to minimize long-term disturbances and maximize adaptability.
\end{axiom}

\subsubsection{III. System Dynamics and Conservation}

\begin{axiom}[Conservation of Learning Energy]
	Total learning energy should remain conserved, with adversarial effects being redistributed rather than lost.
\end{axiom}

\begin{axiom}[Law of Momentum Distribution]
	Momentum can be redistributed within a system in a controlled way to yield strategic advantage, while total momentum is conserved.
\end{axiom}

\begin{axiom}[Law of Controlled Instability]
	Rather than enforcing rigid stability, allowing a degree of controlled instability can help overcome local optima and promote dynamic adaptation.
\end{axiom}

\subsubsection{IV. Multi-agent Cooperation}

\begin{axiom}[Law of Cooperative Redirection]
	In a multi-agent system, optimal redirection is achieved through cooperative adjustments, where each entity aligns its redirection strategy to enhance overall system stability.
\end{axiom}

\begin{axiom}[Law of Strategic Redirection]
	For every disturbance, there exists an optimal redirection strategy that neutralizes its negative impact and enhances system adaptability.
\end{axiom}

\subsection{Computational complexity} \label{appendix:computationalcomplexity}
While AERO introduces additional computation per step due to adversarial signal estimation, its design emphasizes local, first-order approximations, avoiding full second-order cost.

\subsection*{Overhead Sources}
\begin{itemize}
	\item One forward + one backward pass to compute $\nabla_{\text{adv}} L$
	\item Cosine similarity and scalar blending
\end{itemize}

\subsection*{Complexity per Iteration}
\begin{itemize}
	\item \textbf{Time:} $\mathcal{O}(2B)$ \hfill (2x the standard gradient cost per batch $B$)
	\item \textbf{Space:} Slightly increased due to storing adversarial perturbations and alignment metrics
\end{itemize}

However, this cost is offset by faster convergence in high-resistance landscapes and improved generalization due to richer supervision from adversarial cues.

\begin{table}[h!]
	\centering
	\caption{Summary}
	\begin{tabular}{|l|c|c|}
		\hline
		\textbf{Component} & \textbf{Traditional SGD/Adam} & \textbf{AERO Optimizer} \\
		\hline
		Gradient Type & Natural only & Natural + redirected adversarial \\
		\hline
		Update Direction & Straight gradient & Redirection vector with adaptive angle \\
		\hline
		Learning Rate & Static/decayed & Adaptively modulated via effort $\gamma_t$ \\
		\hline
		Response to Adversaries & Defensive (minimize effect) & Constructive (learn from perturbations) \\
		\hline
		Complexity & $\mathcal{O}(B)$ & $\mathcal{O}(2B)$, with faster convergence \\
		\hline
	\end{tabular}
\end{table}

\subsection{Pseudocode for AERO Framework}

The AERO framework can be implemented in an algorithmic fashion for practical applications. Below is a simplified pseudocode that outlines the core steps involved in an AERO-based disturbance redirection process.


\begin{lstlisting}[caption={AERO System Update Routine}]
	# Initialize system parameters
	disturbance_force = get_disturbance_input()
	system_state = initialize_system()
	
	# Step 1: Compute optimal redirection based on disturbance
	optimal_redirection_angle = calculate_optimal_redirection(disturbance_force)
	
	# Step 2: Apply redirection transformation
	redirected_force = apply_redirection_transform(disturbance_force, optimal_redirection_angle)
	
	# Step 3: Monitor system adaptation
	adaptation_rate = calculate_adaptation_rate(system_state, redirected_force)
	
	# Step 4: Apply adaptive response based on timing
	timing_delay = compute_optimal_timing(adaptation_rate)
	wait(timing_delay)
	
	# Step 5: Update system state
	system_state = update_system_state(system_state, redirected_force)
	
	# Step 6: Perform momentum redistribution if necessary
	if needs_momentum_redistribution(system_state):
	redistributed_momentum = redistribute_momentum(system_state)
	
	# Step 7: Return final system state and energy efficiency
	final_state, energy_efficiency = evaluate_system_state(system_state)
	return final_state, energy_efficiency
	
	# Helper functions
	def get_disturbance_input():
	# Fetch disturbance from the environment
	return disturbance_force
	
	def calculate_optimal_redirection(disturbance):
	# Apply the Law of Optimal Redirection
	return optimal_angle
	
	def apply_redirection_transform(disturbance, angle):
	# Apply the redirection transformation
	return redirected_force
	
	def calculate_adaptation_rate(state, redirected_force):
	# Measure how quickly the system adapts
	return adaptation_rate
	
	def compute_optimal_timing(adaptation_rate):
	# Determine the best timing to apply redirection
	return optimal_timing
	
	def update_system_state(state, redirected_force):
	# Update the system state based on new redirection
	return new_state
	
	def needs_momentum_redistribution(state):
	# Check if redistribution of momentum is required
	return condition_met
	
	def redistribute_momentum(state):
	# Redistribute momentum across system components
	return redistributed_momentum
	
	def evaluate_system_state(state):
	# Evaluate the system's final state and energy efficiency
	return state, efficiency
\end{lstlisting}
\newpage
\end{document}